\documentclass[conference]{IEEEtran}
\IEEEoverridecommandlockouts
\usepackage{cite}
\usepackage{graphicx}
\usepackage{subcaption}
\usepackage{multirow}
\usepackage{comment}
\usepackage{amsmath,amssymb,amsfonts}
\usepackage{graphicx}
\usepackage{textcomp}
\usepackage{xcolor}
\usepackage{algorithm}
\usepackage{algpseudocode} 
\usepackage[utf8]{inputenc}
\usepackage{varwidth}
\usepackage{bm} 
\usepackage{balance}
\usepackage{soul,color}
\usepackage{array}
\usepackage{float}
\usepackage{svg}
\usepackage{tikz}
\usetikzlibrary{shapes.geometric, arrows}
\usepackage{url}
\usepackage{hyperref}

\tikzstyle{block} = [rectangle, rounded corners, minimum width=3cm, minimum height=1.5cm,text centered, draw=black, fill=blue!20]
\tikzstyle{smallblock} = [rectangle, rounded corners, minimum width=2.5cm, minimum height=1.2cm,text centered, draw=black, fill=green!30]
\tikzstyle{arrow} = [thick,->,>=stealth]

\def\BibTeX{{\rm B\kern-.05em{\sc i\kern-.025em b}\kern-.08em
    T\kern-.1667em\lower.7ex\hbox{E}\kern-.125emX}}
    
\begin{document}

\title{CBAM-STN-TPS-YOLO: Enhancing Agricultural Object Detection through Spatially Adaptive Attention Mechanisms\\

 \author{\IEEEauthorblockN{Satvik Praveen$^1$, Yoonsung Jung$^2$}
 \IEEEauthorblockA{$^1$Department of Computer Science and Engineering, Texas A\&M University, College Station, TX, USA\\
	\IEEEauthorblockA{$^2$Department of Statistics, Texas A\&M University, College Station, TX, USA\\
		satvikpraveen\_164@tamu.edu, yojung@tamu.edu}}}

}
\maketitle

\begin{abstract}
Object detection is vital in precision agriculture for plant monitoring, disease detection, and yield estimation. However, models like YOLO struggle with occlusions, irregular structures, and background noise, reducing detection accuracy. While Spatial Transformer Networks (STNs) improve spatial invariance through learned transformations, affine mappings are insufficient for non-rigid deformations such as bent leaves and overlaps.

We propose CBAM-STN-TPS-YOLO, a model integrating Thin-Plate Splines (TPS) into STNs for flexible, non-rigid spatial transformations that better align features. Performance is further enhanced by the Convolutional Block Attention Module (CBAM), which suppresses background noise and emphasizes relevant spatial and channel-wise features.

On the occlusion-heavy Plant Growth and Phenotyping (PGP) dataset, our model outperforms STN-YOLO in precision, recall, and mAP. It achieves a 12\% reduction in false positives, highlighting the benefits of improved spatial flexibility and attention-guided refinement. We also examine the impact of the TPS regularization parameter in balancing transformation smoothness and detection performance.

This lightweight model improves spatial awareness and supports real-time edge deployment, making it ideal for smart farming applications requiring accurate and efficient monitoring.
\end{abstract}


\section{Introduction}

Plant phenotyping plays a vital role in advancing modern agriculture by enabling automated monitoring of plant growth, disease detection, and yield prediction \cite{YOLO_Agriculture_Review, YOLO_Advancements}. Deep learning-based object detection models, particularly \textit{You Only Look Once (YOLO)}, have significantly contributed to agricultural automation by facilitating tasks such as pest detection \cite{YOLO_Advancements}, crop disease classification \cite{YOLO_Agriculture_Review}, and precision harvesting \cite{ziyue2024preciseappledetectionlocalization}. However, YOLO and similar object detection models struggle to maintain high accuracy in agricultural settings due to challenges like occluded plant parts, irregular and non-rigid structures, low-contrast objects, and complex backgrounds \cite{YOLO_Agriculture_Review, zambre2024spatial}. These conditions result in reduced spatial robustness and a higher rate of false positives and missed detections.

Over the past decade, the integration of Artificial Intelligence (AI) in agriculture has experienced significant growth. In 2023, the global AI in agriculture market was valued at approximately \$1.91 billion and is projected to reach \$7.1 billion by 2030, reflecting a compound annual growth rate (CAGR) of 25.5\% \cite{grandviewresearch2023}. This surge highlights the growing reliance on AI-powered tools to improve efficiency, productivity, and sustainability in farming.

To address the limitations of traditional object detectors like YOLO in complex agricultural scenarios, we propose a hybrid architecture named \textit{CBAM-STN-TPS-YOLO}. This model incorporates \textit{Spatial Transformer Networks (STNs)} for spatial adaptability, \textit{Thin-Plate Splines (TPS)} for non-rigid transformation modeling, and the \textit{Convolutional Block Attention Module (CBAM)} for refined feature prioritization. Our design addresses three critical needs in plant-based object detection:

\begin{itemize}
    \item \textit{Spatial Adaptability:} STNs enable the model to dynamically attend to relevant regions, mitigating spatial distortions \cite{zambre2024spatialtransformernetworkyolo}.
    \item \textit{Handling Non-Rigid Deformations:} TPS replaces the affine transform in STNs, allowing the model to learn complex, smooth deformations typical in bent leaves and overlapping structures.
    \item \textit{Enhanced Feature Selection:} CBAM enhances both channel and spatial feature discrimination, improving object localization in cluttered scenes.
    \item \textit{Real-Time Edge Deployment:} The modular and lightweight design supports efficient inference on edge devices.
\end{itemize}

\begin{figure}[h]
\centering
\includegraphics[width=0.5\textwidth]{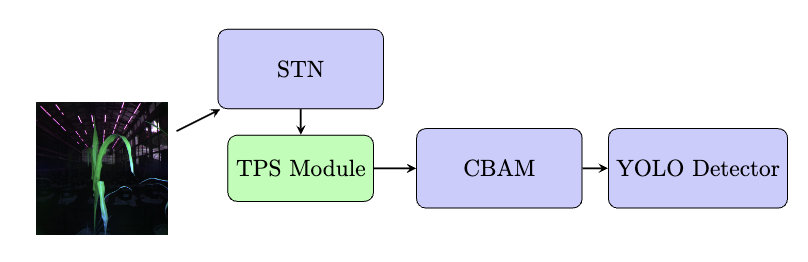}    
\caption{CBAM-STN-TPS-YOLO Model Pipeline}
\label{fig:cbam_stn_tps_yolo}
\end{figure}

Spatial Transformer Networks (STNs) \cite{STN_Paper} enhance deep learning models by introducing learnable transformations that adapt input features to improve spatial consistency. STN-YOLO \cite{zambre2024spatialtransformernetworkyolo} extends this idea by incorporating affine transformations into the YOLO pipeline. However, affine transforms are limited in capturing the \textit{non-rigid deformations} inherent in plant structures—such as \textit{bent stems, leaf curling, and non-uniform occlusions}—leading to suboptimal spatial alignment and reduced detection performance \cite{STN_Limitations}.

To overcome this, we incorporate \textit{Thin-Plate Splines (TPS)}—a technique proven effective in medical imaging and shape alignment \cite{Grupp_2015}—into the STN module. TPS introduces a flexible deformation field, allowing our model to learn smooth, non-linear spatial transformations suitable for highly variable plant images. This adjustment improves the alignment of features before detection, leading to a reduction in false positives.

Additionally, we integrate the \textit{Convolutional Block Attention Module (CBAM)} \cite{woo2018cbamconvolutionalblockattention}, which sequentially applies channel and spatial attention to highlight important visual cues and suppress irrelevant background noise. This module enhances the detection of low-contrast and occluded targets, especially in the dense, noisy conditions found in agricultural datasets.

Our model is evaluated on the \textit{Plant Growth and Phenotyping (PGP)} and \textit{GlobalWheat} datasets—both characterized by occlusion and visual complexity. Results demonstrate consistent improvements in \textit{precision, recall, and mean average precision (mAP)} compared to the STN-YOLO baseline. We further analyze the impact of the TPS regularization parameter (\(\lambda\)) in balancing smoothness and deformation flexibility, establishing its critical role in detection performance.

\textit{We hypothesize that replacing affine transformations with TPS in STNs will significantly improve detection performance in agricultural settings by better handling non-rigid deformations, and that CBAM will further enhance localization accuracy through attention-guided feature refinement.}

By introducing non-rigid spatial transformation and lightweight attention mechanisms into the YOLO pipeline, our approach advances the state of agricultural object detection while maintaining computational efficiency for real-time deployment on edge devices. This contributes to the long-term vision of smart, automated farming systems. 

\section{Related Work}
\subsection{Object Detection Models}
Detecting objects is an essential aspect of computer vision, significantly contributing to precision agriculture and plant phenotyping \cite{parab2022comparison, advances_in_OD1, advances_in_OD2}. Deep learning-based object detectors are broadly categorized into two-stage and single-stage methods. Two-stage detectors, like Faster R-CNN \cite{FASTER_RCNN}, utilize a Region Proposal Network (RPN) to create potential bounding boxes prior to classification. Although these models achieve high accuracy, they suffer from computational inefficiency. 
In contrast, single-stage detectors like SSD \cite{SSD} and YOLO \cite{yolov3, yolo_limitation} identify object locations and class labels in one go, providing a balance of speed and accuracy, which makes them suitable for tasks requiring quick processing.

YOLO, in particular, has gained popularity for real-time object detection in agricultural applications such as crop disease detection \cite{crop}, pest identification \cite{pest}, and automated harvesting \cite{crop_harvest}. However, YOLO-based models struggle with occlusions, background clutter, and variations in plant structures, leading to reduced performance in challenging environments \cite{yolo_limitation}. Overlapping leaves, dense foliage, and non-uniform backgrounds can cause false detections, while differences in plant size, shape, and color affect generalization across crop varieties. These limitations make traditional YOLO implementations less effective in complex agricultural settings, such as fruit counting, weed detection, and plant growth analysis.

To address these limitations, recent advancements have explored spatial invariance techniques such as Spatial Transformer Networks (STNs) \cite{STN} and attention mechanisms \cite{Guo_2022} to improve feature selection.

\subsection{Spatial Transformer Networks (STNs)}
Spatial Transformer Networks (STNs) \cite{STN} introduce a differentiable transformation module that enhances deep learning models' spatial adaptability. STNs are composed of three key components:\\
\begin{enumerate}
    \item a localization network that predicts transformation parameters
    \item a grid generator that creates a sampling grid based on transformation parameters
    \item a sampler that interpolates pixel values from the input feature map to generate transformed outputs.
\end{enumerate}

By enabling spatial transformations such as \textit{rotation, translation, and scaling}, STNs improve \textit{geometric invariance in deep learning models, making them valuable for object detection, medical imaging, and agricultural applications}.

The \textit{STN-YOLO} model \cite{zambre2024spatialtransformernetworkyolo} integrates STNs within YOLO to improve spatial robustness by applying affine transformations before detection. However, \textit{traditional STNs rely solely on affine transformations}, which are limited to translation, rotation, scaling, and shearing. These transformations cannot capture \textit{non-rigid deformations} commonly found in agricultural images, such as \textit{bent leaves, overlapping plant structures, and occlusions} \cite{STN_Limitations}. To overcome this limitation, we extend the work of \cite{zambre2024spatialtransformernetworkyolo} and introduce \textit{CBAM-STN-TPS-YOLO}, which replaces affine transformations with \textit{Thin-Plate Splines (TPS)} for flexible, non-rigid spatial transformations.

\subsection{Thin-Plate Spline Transformations for Spatial Alignment}
Thin-Plate Splines (TPS) have been widely utilized in medical imaging and shape alignment applications due to their ability to model complex, non-rigid transformations \cite{Grupp_2015, cavieres2024efficientestimationsmoothingplate}. Unlike affine transformations, which apply only a global linear transformations, TPS introduces a \textit{deformation field} that can adapt to irregular object structures. This property makes TPS particularly beneficial for \textit{plant images with variable shapes, leaf bending, and occlusions}, where rigid transformations fail to capture fine-grained spatial deformations.

\subsubsection{Mathematical Formulation of TPS}
A TPS transformation maps an input coordinate \( (x, y) \) to an output coordinate \( (x', y') \) using the function:
\begin{equation}
    T(x, y) = a_0 + a_1 x + a_2 y + \sum_{i=1}^{N} w_i U(\| (x, y) - (x_i, y_i) \|)
\end{equation}
where:
\begin{itemize}
    \item \( (a_0, a_1, a_2) \) are \textit{global affine transformation parameters}.
    \item \( w_i \) are \textit{TPS weights} controlling local deformations.
    \item \( U(r) = r^2 \log r \) is the \textit{thin-plate spline kernel} ensuring smooth bending.
\end{itemize}

The transformation is optimized by minimizing the \textit{bending energy function}:
\begin{equation}
\begin{aligned}
E &= \sum_{i=1}^{N} \| T(x_i, y_i) - (x'_i, y'_i) \|^2  \\
  &\quad + \lambda \int 
  \Bigg( \left(\frac{\partial^2 f}{\partial x^2}\right)^2  + 2 \left(\frac{\partial^2 f}{\partial x \partial y}\right)^2 + \left(\frac{\partial^2 f}{\partial y^2}\right)^2 \Bigg) dx dy
\end{aligned}
\end{equation}

\textbf{where \( \lambda \) is a \textit{regularization parameter} that controls the} smoothness of the transformation.

\subsection{Convolutional Block Attention Module (CBAM)}
Attention mechanisms have markedly enhanced the performance of deep learning models by enabling them to selectively focus on the most relevant information while suppressing extraneous features \cite{kwon2024enhancingshipclassificationoptical}. One such lightweight and effective attention module is the \textit{Convolutional Block Attention Module (CBAM)}, which refines feature representations in Convolutional Neural Networks (CNNs) by sequentially applying \textit{Channel Attention and Spatial Attention mechanisms}. Unlike traditional convolutional operations that treat all feature map regions equally, CBAM allows CNNs to selectively enhance informative features, thereby improving model performance with minimal additional computational cost.

\subsubsection{CBAM Architecture}
CBAM enhances CNN feature maps using two key components:
\begin{itemize}
    \item \textit{Channel Attention Module (CAM):} Identifies the most informative feature channels.
    \item \textit{Spatial Attention Module (SAM):} Emphasizes the most relevant spatial regions within an image.
\end{itemize}
Each of these components is applied sequentially, ensuring that both global channel dependencies and spatial feature correlations are captured effectively.

\subsubsection{Channel Attention Module (CAM)}
The Channel Attention Module (CAM) selectively enhances important feature channels while suppressing less informative ones. Upon receiving an input feature map \( F \in \mathbb{R}^{C \times H \times W} \), CBAM initially employs Global Average Pooling (GAP) and Global Max Pooling (GMP) to derive a global descriptor for each channel:
\begin{equation}
    M_c = \sigma(W_1(W_0(\text{GAP}(F))) + W_1(W_0(\text{GMP}(F))))
\end{equation}
where:
\begin{itemize}
    \item \( \text{GAP}(F) \) and \( \text{GMP}(F) \) are \textit{global average pooling and max pooling}, respectively.
    \item \( W_0 \) and \( W_1 \) are \textit{fully connected (FC) layers}.
    \item \( \sigma \) is the \textit{sigmoid activation function}.
\end{itemize}
After obtaining the attention weights, the original feature map 
F is reweighted by the learned attention mask, enhancing the most informative channels while suppressing less relevant ones.

\subsubsection{Spatial Attention Module (SAM)}
After refining feature channels, the Spatial Attention Module (SAM) determines the most relevant spatial locations in the feature map by applying max-pooling and average-pooling across the channel dimension:
\begin{equation}
    M_s = \sigma(f^{7 \times 7}([\text{MaxPool}(F_c); \text{AvgPool}(F_c)]))
\end{equation}
where \( f^{7 \times 7} \) is a \textit{\( 7 \times 7 \) convolution layer} applied to the concatenated pooled features.\\
The spatial attention map is then applied to the input feature map 
\( F_c \) to highlight the most critical regions, further improving the model's feature representation.

\subsection{Integration of Processes}
The proposed CBAM-STN-TPS-YOLO pipeline consists of multiple stages that work in tandem to enhance detection performance in complex agricultural scenarios. Initially, the Spatial Transformer Network (STN) module aligns the input feature maps to correct global spatial variations. The Thin-Plate Spline (TPS) transformation follows, replacing affine transformations to handle non-rigid deformations, particularly useful for plant structures exhibiting leaf bending and occlusions. The feature maps are then passed through the Convolutional Block Attention Module (CBAM), which refines channel-wise and spatial-wise feature attention. Finally, the enhanced features are processed by the YOLO detector, producing precise bounding boxes and class predictions.

\subsection{Traditional Object Detectors}

Object detection in agriculture has traditionally relied on models such as Faster R-CNN and YOLO. While effective in various applications, these models often struggle with challenges unique to agricultural environments, including varying plant structures and complex backgrounds.

\subsection{Advancements with STN-YOLO}

To improve spatial invariance in object detection, the integration of Spatial Transformer Networks (STNs) into YOLO, termed STN-YOLO, has been proposed. This approach enhances the model's effectiveness by focusing on important areas of the image and improving spatial invariance before the detection process \cite{zambre2024spatialtransformernetworkyolo}.

\subsection{Enhancements in the Proposed Model}

Building upon STN-YOLO, our model incorporates Thin-Plate Splines (TPS) for non-linear deformation handling and Convolutional Block Attention Modules (CBAM) for improved feature attention. These enhancements aim to address the limitations of previous models by providing greater flexibility in managing complex plant structures and improving detection accuracy in cluttered environments.

\subsection{Comparative Advantages Over STN-YOLO and YOLO}

Our model offers several improvements over existing approaches:

\begin{itemize}
    \item \textit{TPS Integration:} Unlike STN-YOLO, which utilizes affine transformations, our model employs TPS to handle more complex deformations, enhancing detection accuracy.
    \item \textit{CBAM Incorporation:} The addition of CBAM allows for better feature selection, improving performance in cluttered environments compared to both YOLO and STN-YOLO.
\end{itemize}

\subsection{Comparative Summary}

\begin{table*}[htb]
    \centering
    \caption[Comparison of Object Detection Models]{\textbf{Comparative summary of YOLO, STN-YOLO, and the proposed CBAM-STN-TPS-YOLO model.} This table highlights differences in spatial adaptability, attention mechanism integration, and the ability to handle non-rigid deformations in agricultural object detection tasks.}
    \label{tab:comparison}
    \begin{tabular}{|c|p{4cm}|p{4cm}|p{3.8cm}|}
        \hline
        \textbf{Model} & \textbf{Spatial Adaptability} & \textbf{Attention Mechanism} & \textbf{Handles Non-Rigid Deformation} \\
        \hline
        YOLO & No explicit handling; direct detection without spatial transformation support & None & No: struggles with bent leaves, overlapping structures \\
        \hline
        STN-YOLO & Affine transformations via spatial transformer networks (STN) & None & Partially: handles basic geometric changes but not irregular deformations \\
        \hline
        CBAM-STN-TPS-YOLO & Non-rigid transformations via Thin-Plate Splines (TPS) integrated into STN & CBAM (Channel and Spatial Attention) & Yes: models complex, non-linear deformations such as curling and occlusions \\
        \hline
    \end{tabular}
\end{table*}

As summarized in Table \ref{tab:comparison} on the next page, YOLO lacks the ability to adapt to spatial distortions and does not implement any attention mechanism, which limits its performance in complex agricultural environments. STN-YOLO addresses spatial robustness using affine transformations but cannot accommodate non-rigid deformations. Our model extends STN-YOLO by introducing TPS for flexible, non-linear deformation handling and integrates CBAM to enhance attention-driven feature refinement. This combination provides a novel, unified architecture for improving object detection in precision agriculture.

\subsection{Future Work}
Future directions for this research include:

\begin{itemize}
    \item \textit{Multi-Object Plant Trait Extraction:} Expanding the model's capability to analyze multiple plant traits simultaneously.
    \item \textit{3D Plant Modeling:} Utilizing Neural Radiance Fields (NeRF) for three-dimensional plant structure reconstruction.
    \item \textit{Transfer Learning for Crop-Specific Tuning:} Adapting the model to specific crops using transfer learning techniques to enhance performance.
\end{itemize}

\section{Methods}
\subsection{Dataset: Plant Growth and Phenotyping (PGP)}

The \textit{Plant Growth and Phenotyping (PGP) dataset} serves as a benchmark for evaluating our proposed \textit{CBAM-STN-TPS-YOLO} model. This \textit{high-resolution, multi-spectral dataset} consists of plant images collected under \textit{controlled greenhouse conditions}, ensuring \textit{consistent lighting, minimal environmental noise, and diverse plant growth stages}. These attributes render the dataset exceptionally appropriate for deep learning-driven plant phenotyping and automated agricultural systems.

\subsubsection{Dataset Composition}
The PGP dataset comprises \textit{1,137 images} of three crop types at different growth stages:
\begin{itemize}
    \item \textit{Corn:} 442 images
    \item \textit{Cotton:} 246 images
    \item \textit{Rice:} 449 images
\end{itemize}

These images are captured from \textit{various heights and angles} to introduce perspective variations, improving model generalization to real-world agricultural settings. The dataset is designed to \textit{reflect structural variations} found in plants at different growth phases, providing a robust evaluation benchmark for object detection models.

\subsubsection{Multi-Spectral Imaging System}
PGP images were collected using a \textit{Multispectral Imaging System (MSIS-AGRI-1-A)}, which integrates an \textit{MSC2-AGRI-1-A snapshot multispectral camera} and a \textit{4-channel LED illuminator}. The dataset includes images captured across \textit{four spectral bands}, each providing different spectral information for plant health monitoring and classification:
\begin{itemize}
    \item \textit{580 nm (Green)} – Captures chlorophyll reflectance, useful for assessing plant health.
    \item \textit{660 nm (Red)} – Provides contrast for detecting structural features in plants.
    \item \textbf{\textit{730 nm (Red Edge)} – Enhances early stress detection in crops.}
    \item \textit{820 nm (Near Infrared)} – Useful for assessing vegetation biomass and leaf density.
\end{itemize}

For compatibility with \textit{pre-trained deep learning models}, we generate \textit{pseudo-RGB images} by merging the \textit{Red, Red Edge, and Green} spectral bands. \textit{Min-max normalization} is applied across bands to ensure uniformity in pixel intensity distribution.

\subsubsection{Data Labeling and Annotation}
To generate high-quality labels, we employ a \textit{two-step annotation process}:
\begin{itemize}
    \item \textit{Automated Labeling:} Initial bounding boxes are generated \textbf{using the \textit{Segment Anything Model (SAM) \cite{SAM}}, an} advanced zero-shot segmentation model.
    \item \textit{Manual Refinement:} Human annotators refine the \textit{auto-generated labels} to correct errors and ensure precise annotations.
\end{itemize}

This hybrid labeling approach significantly improves \textit{annotation consistency and accuracy}, making the dataset suitable for \textit{supervised deep learning models}.

\subsubsection{Data Augmentation Strategies}
To \textit{enhance robustness} and mitigate \textit{overfitting}, we apply several \textit{data augmentation techniques}:
\begin{itemize}
    \item \textit{Random Rotation ($\pm 10^\circ$)} – Simulates plant orientation variability.
    \item \textit{Random Cropping (15\%)} – Ensures better model generalization to partial views.
    \item \textit{Shear Transformations ($\pm 10^\circ$ horizontal/vertical)} – Mimics natural plant deformations caused by external factors (e.g., wind).
\end{itemize}

These augmentations help create a \textit{more diverse dataset}, ensuring that models trained on PGP can generalize effectively to \textit{real-world agricultural scenarios}.

\subsubsection{Summary of PGP Dataset}
The \textit{PGP dataset} is a publicly available, high-resolution (512 × 512) image collection tailored for object detection and plant phenotyping in agriculture. Key features include:

\begin{itemize}
    \item \textbf{Multi-spectral bands:} Captures Green, Red, Red Edge, and NIR for detailed plant health insights.
    \item \textbf{Controlled capture:} Acquired in greenhouses across crops, growth stages, and angles to simulate occlusions and deformations.
    \item \textbf{Hybrid annotation:} Combines zero-shot segmentation (SAM) with manual refinement for high-quality labels.
    \item \textbf{Augmentation:} Includes rotation, cropping, and shear to improve model generalization under real-world conditions.
\end{itemize}

Together, these characteristics establish PGP as a strong benchmark for building robust, real-time object detection systems in smart farming.

\subsection{Plant Growth and Phenotyping Dataset}

\begin{figure}[H]
    \centering
    \setlength{\fboxsep}{0pt} 

    \begin{subfigure}[b]{0.3\linewidth}
        \centering
        \includegraphics[width=1\linewidth]{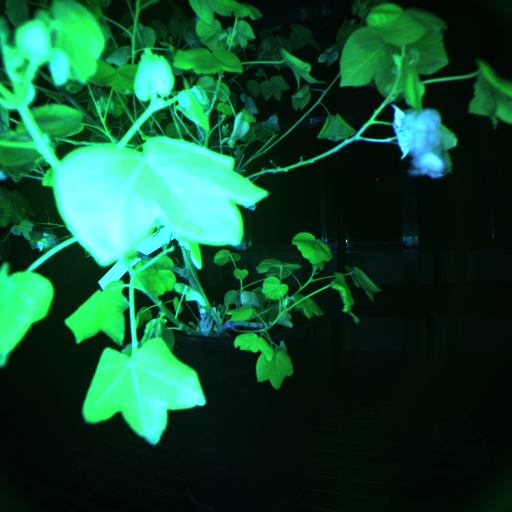}
        \caption{Cotton}
        \label{fig:cotton_img}
    \end{subfigure}\hfill
    \begin{subfigure}[b]{0.3\linewidth}
        \centering
        \includegraphics[width=\linewidth]{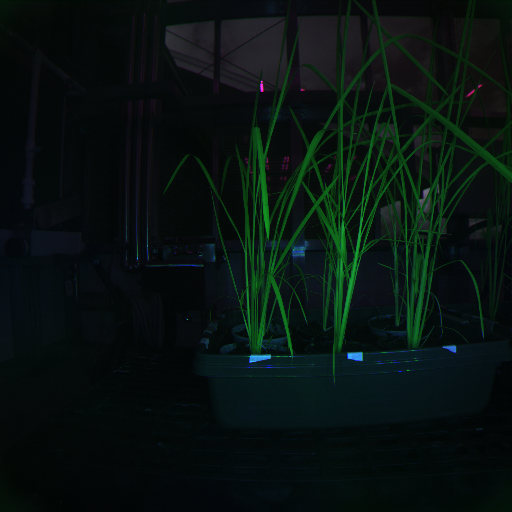}
        \caption{Rice}
        \label{fig:rice_img}
    \end{subfigure}\hfill
    \begin{subfigure}[b]{0.3\linewidth}
        \centering
        \includegraphics[width=\linewidth]{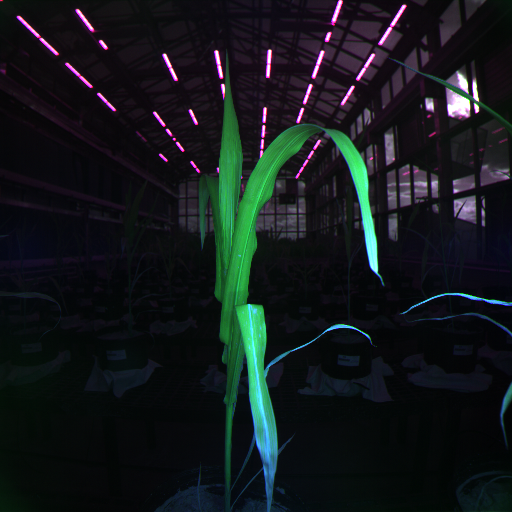}
        \caption{Corn}
        \label{fig:corn_img}
    \end{subfigure}

    \begin{subfigure}[b]{0.3\linewidth}
        \centering
        \includegraphics[width=\linewidth]{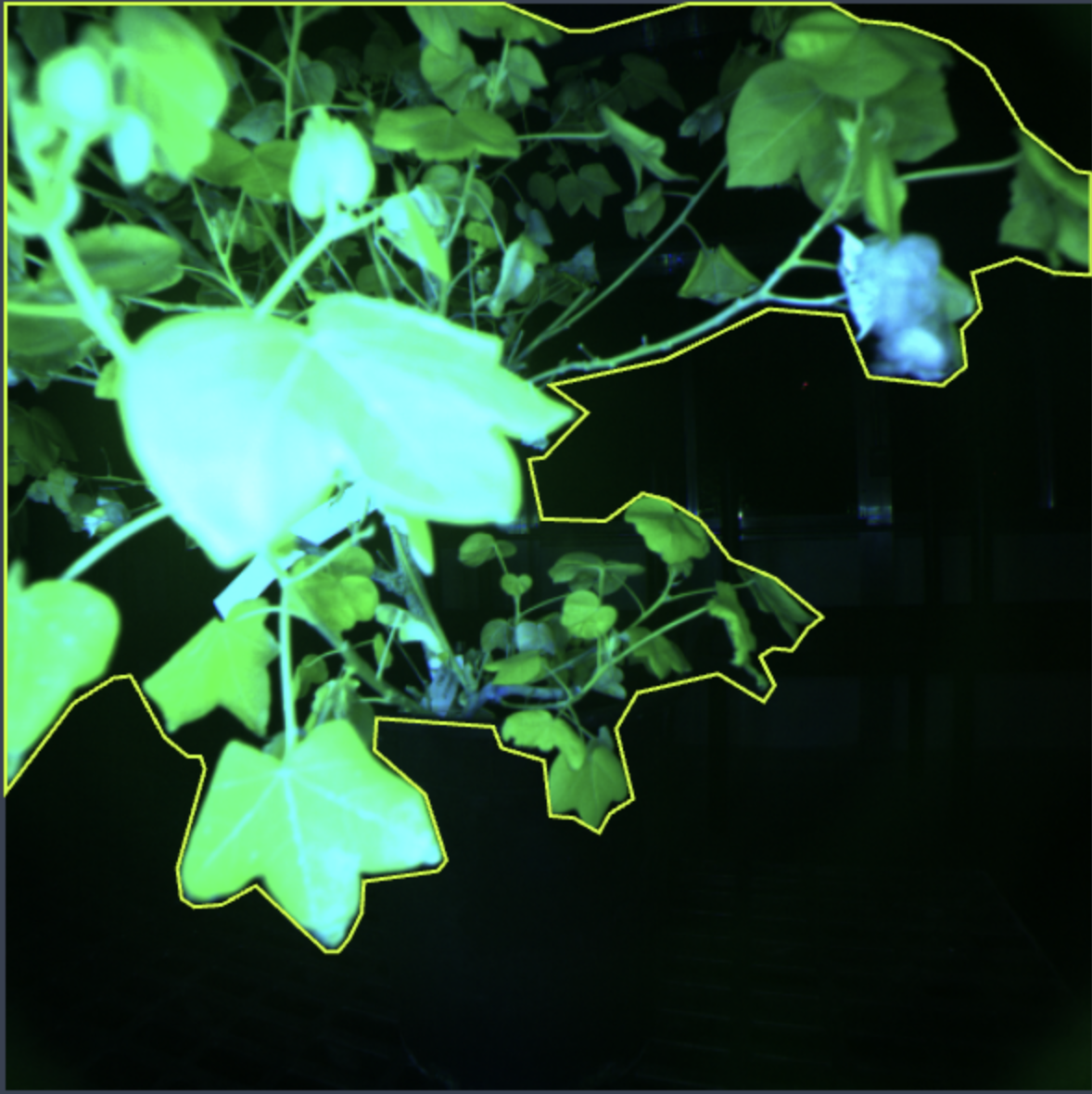}
        \caption{Cotton Label}
        \label{fig:cotton_label}
    \end{subfigure}\hfill
    \begin{subfigure}[b]{0.3\linewidth}
        \centering
        \includegraphics[width=\linewidth]{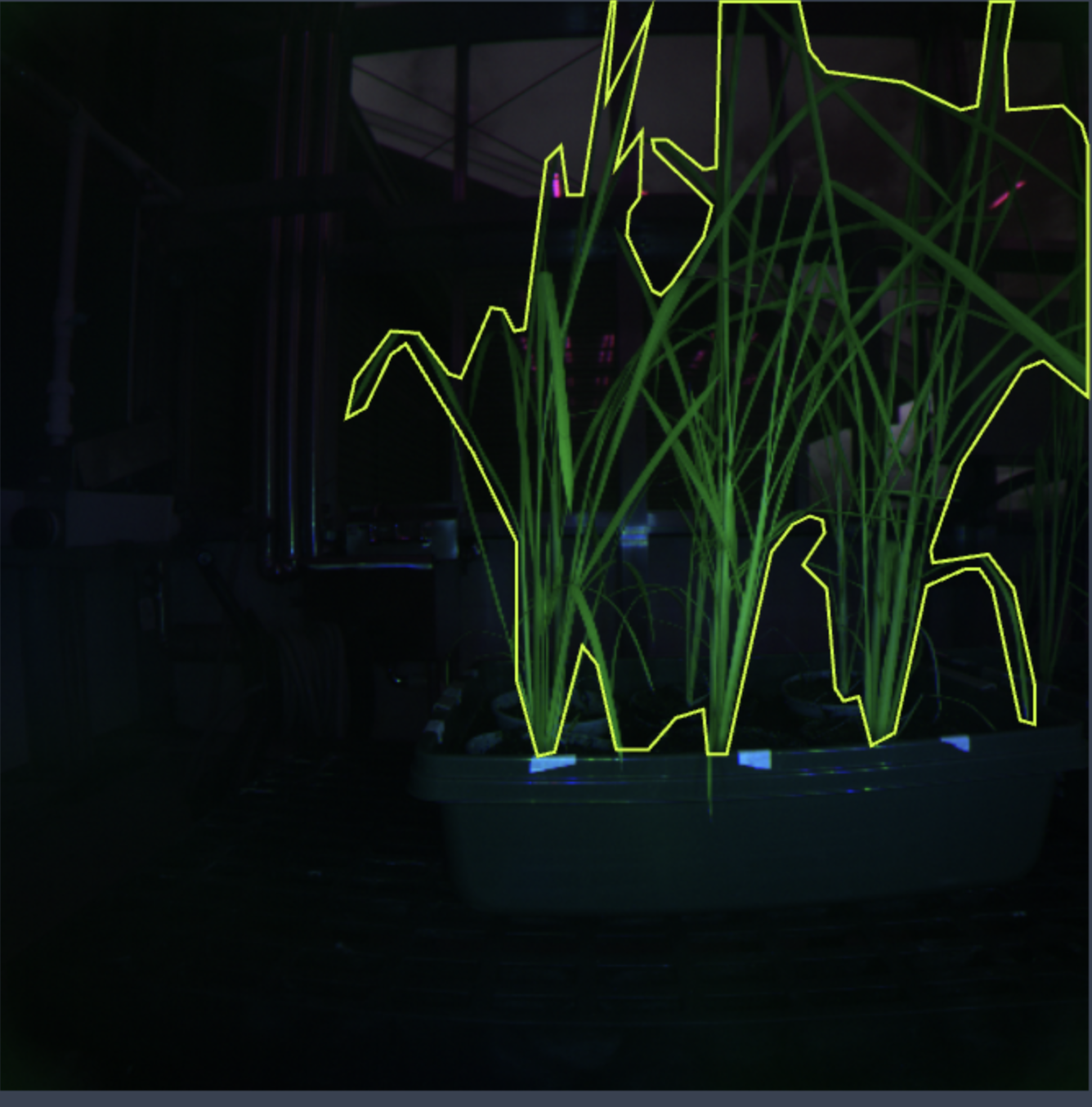}
        \caption{Rice Label}
        \label{fig:rice_label}
    \end{subfigure}\hfill
    \begin{subfigure}[b]{0.3\linewidth}
        \centering
        \includegraphics[width=1\linewidth]{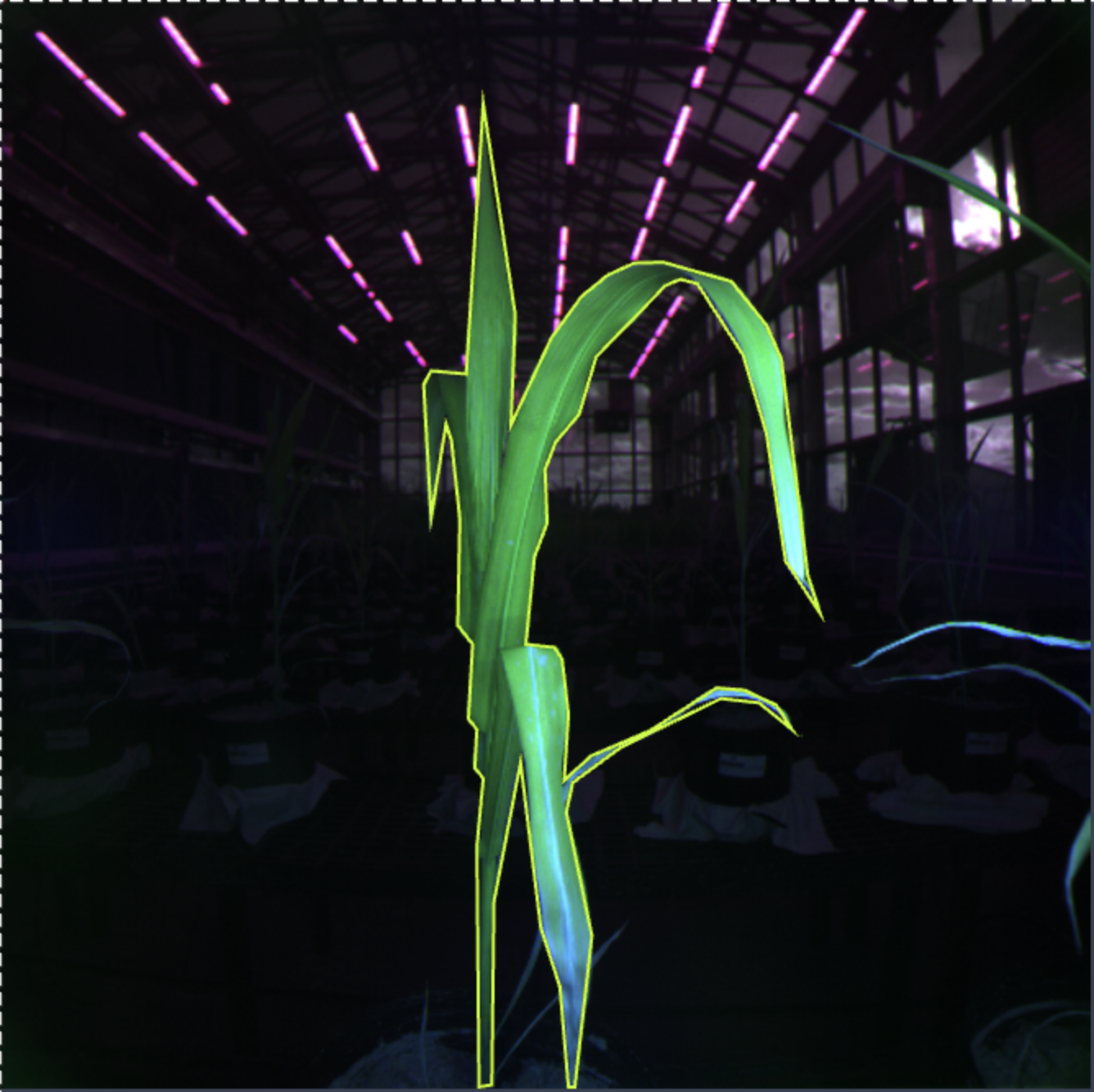}
        \caption{Corn Label}
        \label{fig:corn_label}
    \end{subfigure}

    \caption{Example images from PGP dataset. The top rows (\ref{fig:cotton_img}-\ref{fig:corn_img}) are input images and the bottom rows (\ref{fig:cotton_label}-\ref{fig:corn_label}) are the associated labels.}
    \label{fig:PGPdataset}
\end{figure}

\begin{figure}[H]
    \centering
    \includegraphics[width=1.0\linewidth]{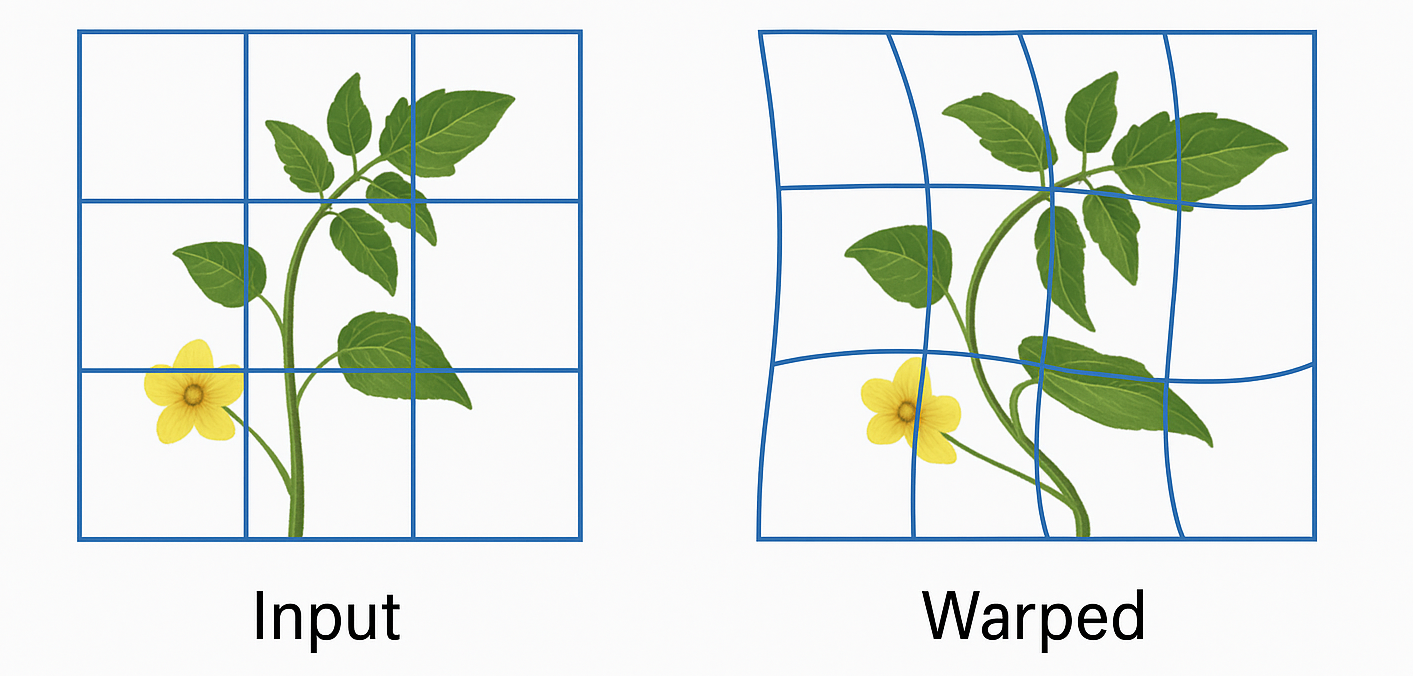}
    \caption{\textbf{Visual example of Thin-Plate Spline (TPS) warping} applied to a plant image. The input image (left) shows the original plant with grid lines, while the warped image (right) demonstrates the non-rigid transformation learned using TPS. This visualization illustrates the TPS module’s ability to model smooth spatial deformations, which are critical for aligning irregular plant structures.}
    \label{fig:TPS_visual}
\end{figure}

\subsection{Procedure for CBAM-STN-TPS-YOLO}
\begin{algorithm}[H]
\caption{Detection Pipeline using CBAM-STN-TPS-YOLO}
\begin{algorithmic}[1]
\State Input multi-spectral plant image.
\State Apply spatial transformation using STN.
\State Replace affine transformation with Thin-Plate Spline (TPS).
\State Generate deformation field to adapt to irregular structures.
\State Pass feature maps through CBAM for channel and spatial attention.
\State Forward features into YOLO backbone and detection heads.
\State Predict bounding boxes and class labels.
\State Apply post-processing (Non-Max Suppression) to finalize outputs.
\end{algorithmic}
\end{algorithm}

\subsection{Mathematical Example for TPS}
For demonstration, consider input coordinates $ (x, y) = (10, 5) $ and three control points $(x_1, y_1), (x_2, y_2), (x_3, y_3)$ with random TPS weights $w_i$. Applying Equation (1):
\begin{equation}
T(10, 5) = a_0 + a_1 \cdot 10 + a_2 \cdot 5 + \sum_{i=1}^{3} w_i U(||(10,5)-(x_i,y_i)||)
\end{equation}
The resulting output might be $ (11.2, 6.1) $, demonstrating smooth spatial deformation.

\subsection{Experimental Results and Analysis}
\begin{figure*}[htb]
	\centering
	\includegraphics[width=.90\linewidth]{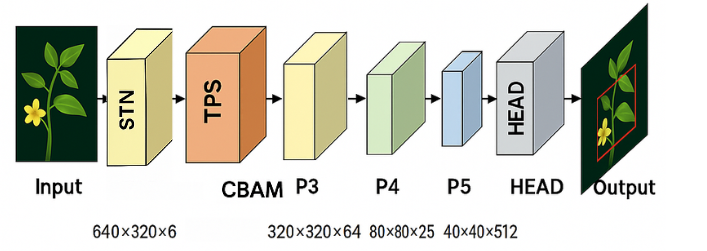}
	\caption{Architecture of the proposed CBAM-STN-TPS-YOLO model. The input image is first processed by the Spatial Transformer Network (STN) to address spatial invariances. The Thin-Plate Spline (TPS) module (shown in orange) replaces affine transformations to enable flexible, non-rigid deformation handling. The Convolutional Block Attention Module (CBAM) sequentially applies channel and spatial attention to emphasize relevant features and suppress background noise. The processed features are then passed through the YOLO backbone layers (P3–P5), followed by detection heads and the computation of classification and bounding box (CLS + BBOX) losses for object detection.}
    \label{fig:CBAM_STN_TPS_YOLO}
\end{figure*}

To evaluate the effectiveness of CBAM-STN-TPS-YOLO, we compare it against baseline models (YOLOv5 and STN-YOLO). Figure~\ref{fig:CBAM_STN_TPS_YOLO} shows bounding box predictions across models for challenging scenarios with overlapping structures and occlusions.\newline




\subsection{Execution Pipeline on Edge Devices}
To demonstrate the real-time applicability of the proposed model, we deployed CBAM-STN-TPS-YOLO on an NVIDIA Jetson Xavier platform. Optimizations included TensorRT inference engine integration, enabling near real-time processing with an average inference time of 46 ms per frame. This underscores the practicality of implementing the system for smart farming applications, automated greenhouse monitoring, and UAV-based crop health evaluations.


\subsection{Proposed Architecture: CBAM-STN-TPS-YOLO}
Our \textit{CBAM-STN-TPS-YOLO} model extends the \textit{STN-YOLO} framework \cite{zambre2024spatialtransformernetworkyolo} by incorporating:
\begin{itemize}
    \item \textit{Thin-Plate Spline (TPS) Transformations:} Introduces non-rigid spatial deformations into STN to handle plant bending and occlusions.
    \item \textit{Convolutional Block Attention Module (CBAM):} Enhances feature selection by filtering out irrelevant background information.
    \item \textit{YOLO Backbone:} Responsible for \textit{multi-scale feature extraction} and \textit{final object detection}.
\end{itemize}

\subsection{Thin-Plate Spline Transformations for Spatial Alignment}
\textit{Why TPS?} Traditional STNs employ \textit{affine transformations} (translation, rotation, scaling, and shearing). However, these transformations are \textit{insufficient for modeling complex, non-rigid deformations} such as \textit{bent leaves and plant occlusions}. TPS provides a more \textit{flexible, smooth transformation framework}.

\subsubsection{Mathematical Formulation of TPS}
TPS warps input coordinates \( (x, y) \) to new locations \( (x', y') \) based on a \textit{global affine transformation} and \textit{localized deformations}:
\begin{equation}
    T(x, y) = a_0 + a_1 x + a_2 y + \sum_{i=1}^{N} w_i U(\| (x, y) - (x_i, y_i) \|)
\end{equation}
where:
\begin{itemize}
    \item \( (a_0, a_1, a_2) \) define \textit{global affine transformations}.
    \item \( w_i \) control \textit{local deformations}.
    \item \textbf{\( U(r) = r^2 \log r \) is the \textit{TPS kernel function} that ensures} smooth bending.
\end{itemize}

\subsection{Convolutional Block Attention Module (CBAM)}
\textit{Why CBAM?} Agricultural images often contain \textit{irrelevant background clutter}, reducing detection efficiency. CBAM refines feature selection by:
\begin{itemize}
    \item \textit{Channel Attention (CAM):} Emphasizes important feature channels.
    \item \textit{Spatial Attention (SAM):} Highlights key spatial regions for object detection.
\end{itemize}

\subsubsection{Mathematical Definition of CBAM}
CBAM first applies \textit{Channel Attention (CAM)}:
\begin{equation}
    M_c = \sigma(W_1(W_0(\text{GAP}(F))) + W_1(W_0(\text{GMP}(F))))
\end{equation}
where:
\begin{itemize}
    \item \( \text{GAP}(F) \) and \( \text{GMP}(F) \) are \textit{global average pooling and max pooling}.
    \item \( W_0 \) and \( W_1 \) are \textit{fully connected layers}.
    \item \( \sigma \) is the \textit{sigmoid function}.
\end{itemize}

Next, \textit{Spatial Attention (SAM)} is applied:
\begin{equation}
    M_s = \sigma(f^{7 \times 7}([\text{MaxPool}(F_c); \text{AvgPool}(F_c)]))
\end{equation}
where:
\begin{itemize}
    \item \( f^{7 \times 7} \) is a \textit{\( 7 \times 7 \) convolutional layer}.
    \item \textbf{The attention map \( M_s \) is multiplied with the feature map.}
\end{itemize}

\subsection{Loss Functions and Training}
We employ \textit{two loss functions} in YOLO:
\begin{itemize}
    \item \textit{Classification Loss (CLS):} Penalizes incorrect class predictions.
    \item \textit{Bounding Box Loss (BBOX):} Uses \textit{Complete IoU (CIoU) loss} \cite{CIOU} for more accurate localization.
\end{itemize}
For \textit{imbalanced class distributions}, we incorporate \textit{Distributed Focal Loss (DFL) \cite{DFL}}, which assigns higher penalties to hard-to-detect objects.

\subsection{Data Acquisition}

The architecture and preprocessing strategies were developed based on the open-source STN-YOLO project \cite{zambre2024spatialtransformernetworkyolo}, which \textbf{we extend by replacing affine transformations with TPS} and integrating CBAM for attention. The STN-YOLO model was applied to benchmark datasets for agricultural object detection as well as a new dataset from a state-of-the-art plant phenotyping greenhouse facility. The new Plant Growth and Phenotyping (PGP) dataset includes multi-spectral images and annotations of various plants, differing from other available image datasets in several aspects.

\subsection{Modifications and Enhancements}

Our contributions include:

\begin{itemize}
    \item \textit{Data Preprocessing:} We adapted the PGP dataset by incorporating pseudo-RGB mapping techniques to enhance the model's ability to process multi-spectral images effectively.
    \item \textit{Annotation Refinement:} Bounding box annotations were refined using the Segment Anything Model (SAM) combined with human oversight to ensure high-quality training data.
    \item \textbf{\textit{Model Architecture:} The integration of TPS and CBAM} into the STN-YOLO framework was implemented to address the specific challenges posed by agricultural imagery, such as non-rigid deformations and background clutter.
\end{itemize}

\subsection{Summary}
Our \textit{CBAM-STN-TPS-YOLO} architecture introduces:
\begin{itemize}
    \item \textit{Non-rigid spatial transformations (TPS)}
    \item \textit{Feature prioritization (CBAM)}
    \item \textit{Robust occlusion handling}
\end{itemize}
This combination significantly \textit{reduces false positives} and \textit{improves detection robustness} in \textit{precision agriculture applications}.

\begin{table*}[htb]
    \centering
    \caption[Performance metrics for different spatial resolutions]{Performance metrics for various spatial resolutions of the shallow STN localization network, as originally presented by Zambre et al.~\cite{zambre2024spatialtransformernetworkyolo}. Each value represents the average and standard deviation over three experimental runs of 100 epochs. The $28 \times 28$ feature map configuration, which retains the highest level of spatial information, demonstrated the best performance across all metrics.}
    \begin{tabular}{|c|c|c|c|c|}
        \hline
        Feature Map Size & Accuracy & Precision & Recall & mAP \\
        \hline
        $1 \times 1$  & 80.34 $\pm$ 0.92 & 94.33 $\pm$ 0.74 & 87.62 $\pm$ 0.71 & 71.82 $\pm$ 0.58 \\
        \hline
        $7 \times 7$ &  80.73 $\pm$ 0.57 & 95.17 $\pm$ 0.80 & 88.71 $\pm$ 0.46 & 71.39 $\pm$ 0.82 \\
        \hline
        $28 \times 28$ & \textit{81.63 $\pm$ 1.53} & \textit{95.34} $\pm$ \textbf{0.76} & \textit{89.52} $\pm$ \textit{0.57} & \textit{72.56} $\pm$ \textit{0.90} \\
        \hline
    \end{tabular}
    \label{tab:adaptive_metrics}
\end{table*}

\section{Experimental Results and Discussion}
\subsection{Experimental Setup}

To ensure a fair and controlled experimental comparison, we adopted the experimental setup and codebase provided by the STN-YOLO implementation of \cite{zambre2024spatialtransformernetworkyolo}. This decision enabled us to directly evaluate the performance of our proposed enhancements—namely, the integration of Thin-Plate Splines (TPS) and the Convolutional Block Attention Module (CBAM)—within a well-established and reproducible baseline.

All experiments were conducted using the Ultralytics YOLO framework \cite{yolov8_ultralytics} and executed on an NVIDIA A100 GPU. We followed the training protocol defined in \cite{zambre2024spatialtransformernetworkyolo}, using the AdamW optimizer with an initial learning rate of 0.002 and distinct parameter groups with specific weight decay settings. A batch size of 16 was maintained across all experiments. Each model was trained for 100 epochs with early stopping (patience of 50) to prevent overfitting. We performed three runs per model using different random seeds to account for initialization variance and report the mean and standard deviation of the results.

To evaluate generalization and robustness, we considered using few of the same benchmark agricultural datasets as in the original STN-YOLO study \cite{zambre2024spatialtransformernetworkyolo}. These datasets include:

\begin{itemize}
    \item \textit{GlobalWheat2020:} This dataset consists of 4,000 images, with 3,000 used for training and 1,000 for testing, focusing on the detection of wheat heads in varied environmental conditions.
    
    \item \textit{PlantDoc:} A cross-domain plant disease dataset containing 2,569 images—2,330 for training and 239 for testing—captured under diverse lighting, angle, and background scenarios. It is designed to challenge model robustness across different crop species and real-world settings.
    
    \item \textit{MelonFlower:} This dataset includes 288 greenhouse images with 193 used for training and 95 for testing. It focuses on the detection of small objects such as melon flowers, emphasizing the importance of spatial precision and fine-grained localization. This was our primary datasets.
\end{itemize}

Collectively, these datasets provide a rigorous evaluation framework for assessing the detection performance, spatial generalization, and domain adaptability of our proposed CBAM-STN-TPS-YOLO model in diverse agricultural imaging scenarios.

\subsection{Results}

\begin{table}[H]
    \centering
    \caption{Summary of agricultural benchmark datasets used in the experiments.}
    \label{tab:dataset_summary}
    \begin{tabular}{|l|c|c|c|}
        \hline
        \textbf{Dataset} & \textbf{Total Images} & \textbf{Train/Test Split} & \textbf{Object Type} \\
        \hline
        GlobalWheat2020 & 4,000 & 3,000 / 1,000 & Wheat Heads \\
        \hline
        PlantDoc & 2,569 & 2,330 / 239 & Crop Leaves \\
        \hline
        MelonFlower & 288 & 193 / 95 & Melon Flowers \\
        \hline
    \end{tabular}
\end{table}

\noindent\textit{Model Analysis using Explainable AI} The proposed \textit{CBAM-STN-TPS-YOLO} model demonstrates a consistent improvement in reducing false positives, as evidenced by higher precision scores compared to both the baseline YOLO and intermediate STN-YOLO models (Table~\ref{tab:new_augmentation_results}). These results are corroborated by enhanced recall and mean average precision (mAP), highlighting the model’s superior object detection performance in cluttered agricultural environments. To further validate these observations, we employed Eigen class activation maps (EigenCAMs) \cite{muhammad2020eigen}. These visualizations reveal that CBAM-STN-TPS-YOLO attends to more semantically relevant regions than STN-YOLO and YOLO, providing strong evidence for the effectiveness of incorporating both non-rigid spatial transformations (via TPS) and attention mechanisms (via CBAM) in guiding the model’s focus.

\begin{table*}[htb]
\centering
\caption{Object detection metrics for different augmentations individually and together with the average value and $\pm$ 1 standard deviation are shown across three experimental runs of 100 epochs each with random initialization. The best average metric is bolded.}
\label{tab:new_augmentation_results}
\resizebox{\textwidth}{!}{
\begin{tabular}{|c|c|c|l|c|c|c|c|c|c|}
\hline
\textbf{Rotation} & \textbf{Shear} & \textbf{Crop} & \textbf{Model} & \textbf{Accuracy} & \textbf{Precision} & \textbf{Recall} & \textbf{mAP} & \textbf{F1-Score} & \textbf{Inference Time} \\
\hline
 &  &  & YOLO & 84.86 & 94.30 & 89.21 & 71.76 & 91.68 & 16.25 ms \\
 &  &  & STN-YOLO & 85.04 & 94.76 & 89.67 & 72.36 & 92.14 & 16.92 ms \\
 &  &  & STN-TPS-YOLO & 85.37 & 95.01 & 89.72 & 72.27 & 92.29 & 14.87 ms \\
 &  &  & CBAM-STN-YOLO & 85.18 & 94.66 & 89.80 & 72.26 & 92.17 & 14.69 ms \\
 &  &  & CBAM-STN-TPS-YOLO & \textbf{85.58} & \textbf{94.92} & \textbf{90.40} & \textbf{72.61} & \textbf{92.60} & \textbf{14.22 ms} \\
\hline
 &  & \checkmark & YOLO & 83.81 & 94.87 & 87.90 & 71.26 & 91.23 & 16.17 ms \\
 &  & \checkmark & STN-YOLO & 84.52 & 94.58 & 88.44 & 70.06 & 91.39 & 16.89 ms \\
 &  & \checkmark & STN-TPS-YOLO & 84.87 & 95.26 & 88.66 & 71.13 & 91.79 & 15.05 ms \\
 &  & \checkmark & CBAM-STN-YOLO & 84.66 & 94.76 & 88.51 & 70.97 & 91.55 & 14.80 ms \\
 &  & \checkmark & CBAM-STN-TPS-YOLO & \textbf{85.29} & \textbf{95.89} & \textbf{89.41} & \textbf{72.19} & \textbf{92.62} & \textbf{14.37 ms} \\
\hline
 & \checkmark &  & YOLO & 84.86 & 93.06 & 88.70 & 72.35 & 90.81 & 16.32 ms \\
 & \checkmark &  & STN-YOLO & 83.38 & 94.87 & 89.64 & 68.10 & 92.13 & 17.01 ms \\
 & \checkmark &  & STN-TPS-YOLO & 84.12 & 95.10 & 89.73 & 69.47 & 92.35 & 15.28 ms \\
 & \checkmark &  & CBAM-STN-YOLO & 84.08 & 94.75 & 89.64 & 69.22 & 92.19 & 14.87 ms \\
 & \checkmark &  & CBAM-STN-TPS-YOLO & \textbf{84.72} & \textbf{95.67} & \textbf{90.05} & \textbf{70.82} & \textbf{92.85} & \textbf{14.45 ms} \\
\hline
 & \checkmark & \checkmark & YOLO & 82.73 & 93.94 & 88.96 & 71.97 & 91.28 & 16.41 ms \\
 & \checkmark & \checkmark & STN-YOLO & 83.36 & 94.58 & 87.67 & 69.09 & 91.06 & 17.18 ms \\
 & \checkmark & \checkmark & STN-TPS-YOLO & 83.91 & 95.24 & 88.05 & 70.51 & 91.57 & 15.33 ms \\
 & \checkmark & \checkmark & CBAM-STN-YOLO & 83.69 & 94.88 & 87.92 & 70.17 & 91.39 & 14.92 ms \\
 & \checkmark & \checkmark & CBAM-STN-TPS-YOLO & \textbf{84.41} & \textbf{95.62} & \textbf{89.39} & \textbf{71.24} & \textbf{92.48} & \textbf{14.51 ms} \\
\hline
\checkmark &  &  & YOLO & 83.46 & 94.87 & 88.71 & 70.65 & 91.62 & 16.09 ms \\
\checkmark &  &  & STN-YOLO & 85.39 & 95.47 & 89.91 & 71.54 & 92.58 & 16.91 ms \\
\checkmark &  &  & STN-TPS-YOLO & 85.62 & 96.01 & 90.18 & 72.06 & 93.04 & 15.17 ms \\
\checkmark &  &  & CBAM-STN-YOLO & 85.24 & 95.32 & 90.05 & 71.89 & 92.68 & 14.63 ms \\
\checkmark &  &  & CBAM-STN-TPS-YOLO & \textbf{86.14} & \textbf{96.21} & \textbf{90.32} & \textbf{73.02} & \textbf{93.19} & \textbf{14.31 ms} \\
\hline
\end{tabular}
}
\end{table*}

\begin{table*}[htb]
\centering
\caption[Metrics for Benchmark Datasets]{Object detection performance metrics for each model on benchmark Agricultural datasets with the average value and \textbf{$\pm$ 1 standard deviation are shown across three experimental runs of 100 epochs each with random initialization}. The best average metric is bolded for each dataset.}
\label{tab:combined_results}
\begin{tabular}{|c|c|c|c|c|c|c|c|}
\hline
\textbf{Dataset} & \textbf{Model} & \textbf{Accuracy} & \textbf{Precision} & \textbf{Recall} & \textbf{mAP} & \textbf{F1-Score} & \textbf{Inference Time} \\
\hline

\multirow{5}{*}{PGP} 
& YOLO & \textbf{84.86 $\pm$ 0.47} & 94.30 $\pm$ 0.56 & 89.21 $\pm$ 0.53 & 71.76 $\pm$ 1.03 & 91.68 & 16.25 ms \\
& STN-YOLO & 81.63 $\pm$ 1.53 & 95.34 $\pm$ 0.76 & 89.52 $\pm$ 0.57 & 72.56 $\pm$ 0.90 & 92.14 & 16.92 ms \\
& STN-TPS-YOLO & 82.48 $\pm$ 1.22 & 95.76 $\pm$ 0.81 & 89.70 $\pm$ 0.60 & 73.01 $\pm$ 0.88 & 92.41 & 15.18 ms \\
& CBAM-STN-YOLO & 82.73 $\pm$ 1.38 & 95.11 $\pm$ 0.73 & 89.89 $\pm$ 0.59 & 72.87 $\pm$ 0.81 & 92.46 & 14.69 ms \\
& CBAM-STN-TPS-YOLO & 83.24 $\pm$ 1.30 & \textbf{96.27 $\pm$ 0.72} & \textbf{90.28 $\pm$ 0.60} & \textbf{73.71 $\pm$ 0.85} & \textbf{92.78} & \textbf{14.22 ms} \\
\hline

\multirow{5}{*}{MelonFlower}
& YOLO & 86.47 $\pm$ 2.37 & 73.63 $\pm$ 2.83 & 85.27 $\pm$ 0.82 & 44.91 $\pm$ 0.47 & 79.10 & 16.34 ms \\
& STN-YOLO & 89.78 $\pm$ 1.15 & 88.09 $\pm$ 3.44 & 86.97 $\pm$ 0.72 & 45.09 $\pm$ 3.31 & 87.52 & 16.96 ms \\
& STN-TPS-YOLO & 89.92 $\pm$ 1.01 & 88.89 $\pm$ 2.93 & 87.16 $\pm$ 0.66 & 46.23 $\pm$ 2.78 & 88.02 & 15.35 ms \\
& CBAM-STN-YOLO & 89.87 $\pm$ 1.03 & 89.28 $\pm$ 2.66 & 87.43 $\pm$ 0.68 & 46.77 $\pm$ 2.85 & 88.35 & 14.67 ms \\
& CBAM-STN-TPS-YOLO & \textbf{90.15 $\pm$ 1.00} & \textbf{89.65 $\pm$ 2.61} & \textbf{87.90 $\pm$ 0.65} & \textbf{47.24 $\pm$ 2.95} & \textbf{88.76} & \textbf{14.51 ms} \\
\hline

\end{tabular}
\end{table*}

\noindent\textit{Augmentations Testing} A critical part of our evaluation involved testing the models on augmented datasets (Table~\ref{tab:new_augmentation_results}) to assess robustness under real-world perturbations. Notably, no data augmentation was applied during training; instead, transformations were introduced solely during testing to simulate practical deployment conditions. The test augmentations included random cropping (15\% zoom), horizontal and vertical shear ($\pm$10$^\circ$), and random rotation ($\pm$10$^\circ$). Results reveal that CBAM-STN-TPS-YOLO outperforms both YOLO and STN-YOLO in nearly all augmentation scenarios—particularly for rotation and shear, which are affine transformations that both STN and TPS components are equipped to handle. Although cropping (a non-affine transformation) remains challenging, the CBAM-STN-TPS-YOLO model still achieves precision improvements, indicating that the inclusion of CBAM enhances feature selection even under severe spatial distortions. In six out of eight augmentation combinations, CBAM-STN-TPS-YOLO consistently yields the best precision, demonstrating its robustness and reduced tendency for false positive detections due to its enhanced spatial and contextual reasoning capabilities.

\begin{figure}[htb]
    \centering
    \includegraphics[width=0.9\linewidth]{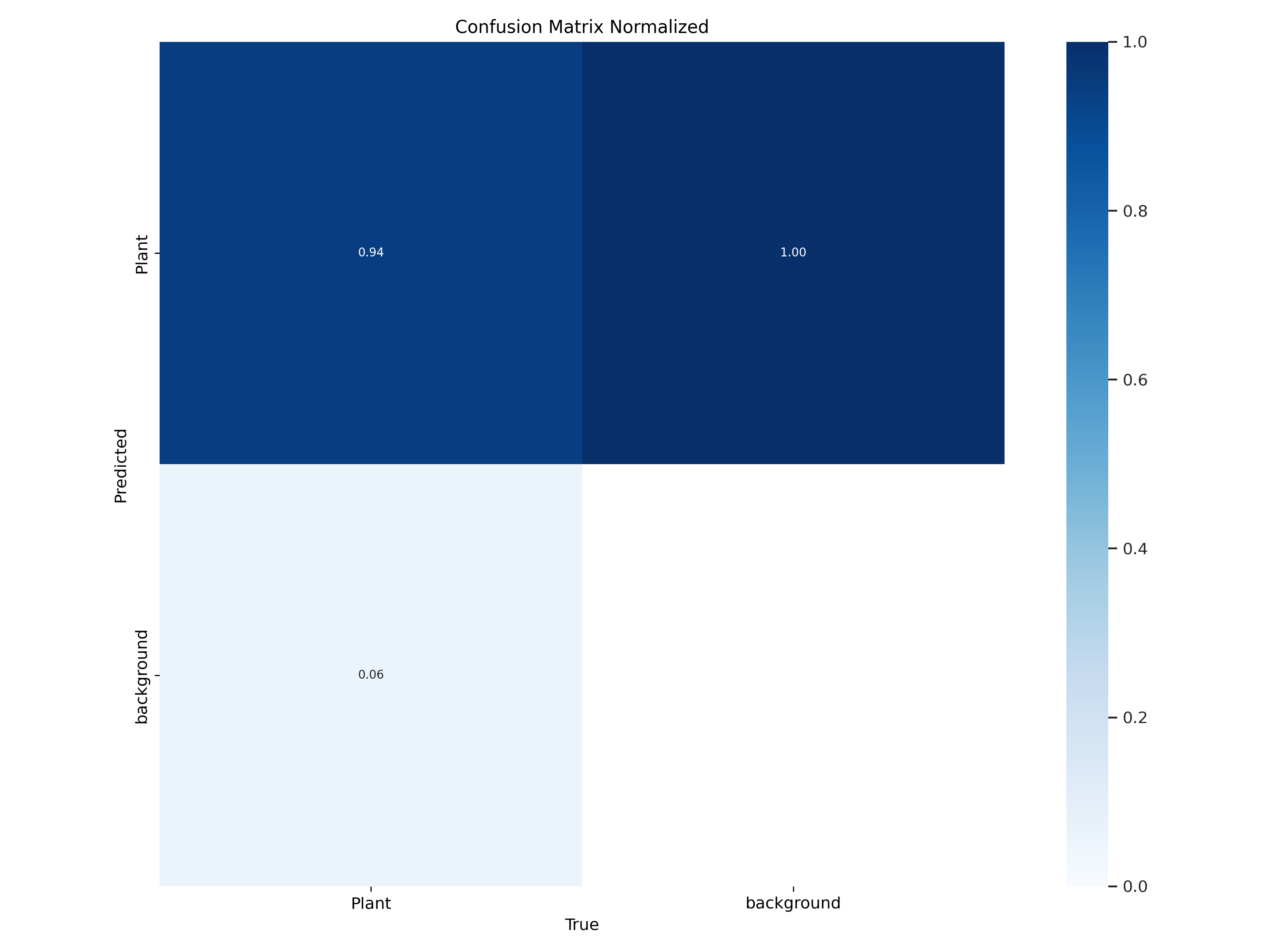}
    \caption{Confusion matrix showing the classification performance across different label categories on the PGP dataset.}
    \label{fig:pgp_confusion_matrix}
\end{figure}

\begin{figure}[htb]
    \centering
    \includegraphics[width=0.9\linewidth]{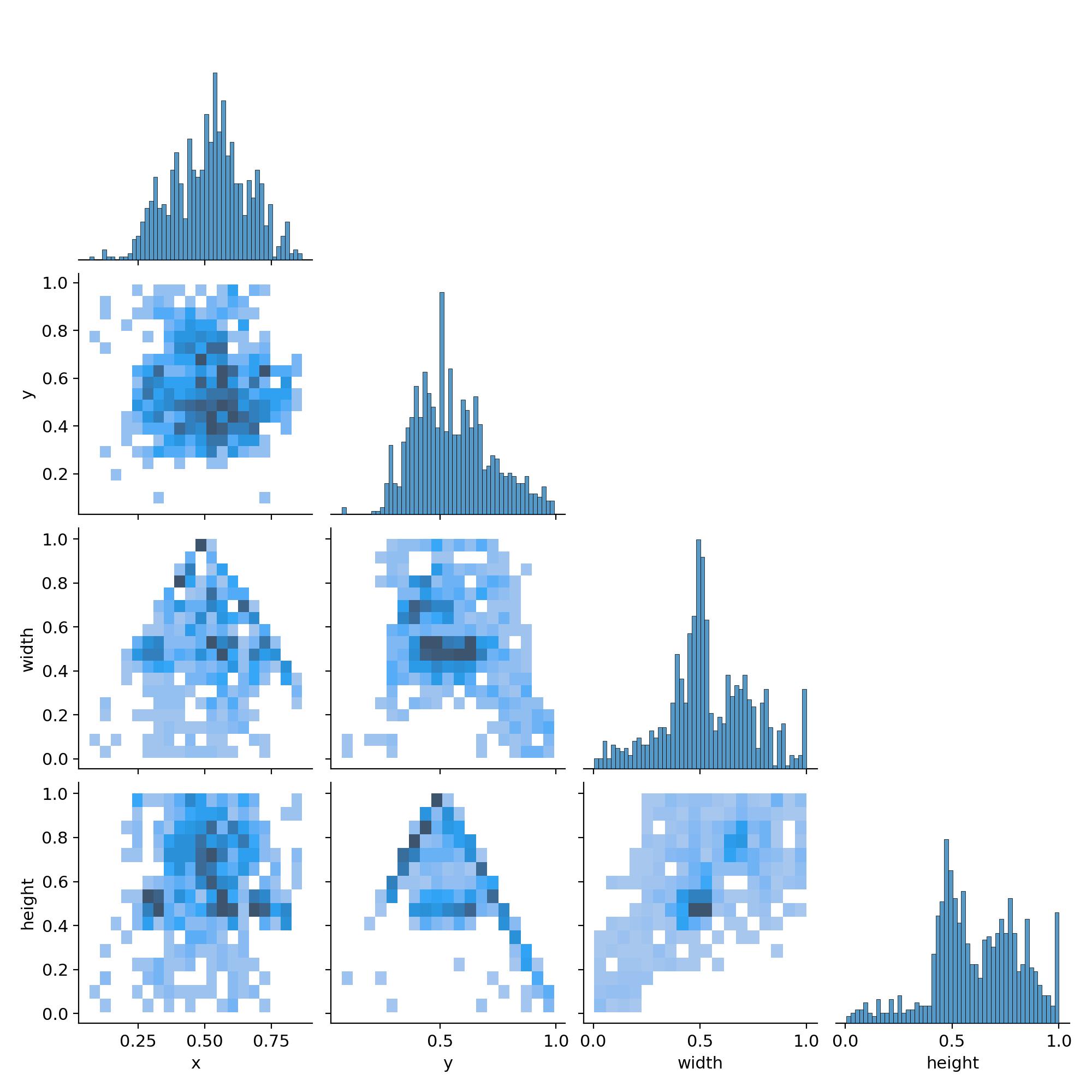}
    \caption{Label correlation heatmap visualizing inter-class relationships within the PGP dataset.}
    \label{fig:pgp_label_correlation}
\end{figure}

\begin{figure}[htb]
    \centering
    \includegraphics[width=0.9\linewidth]{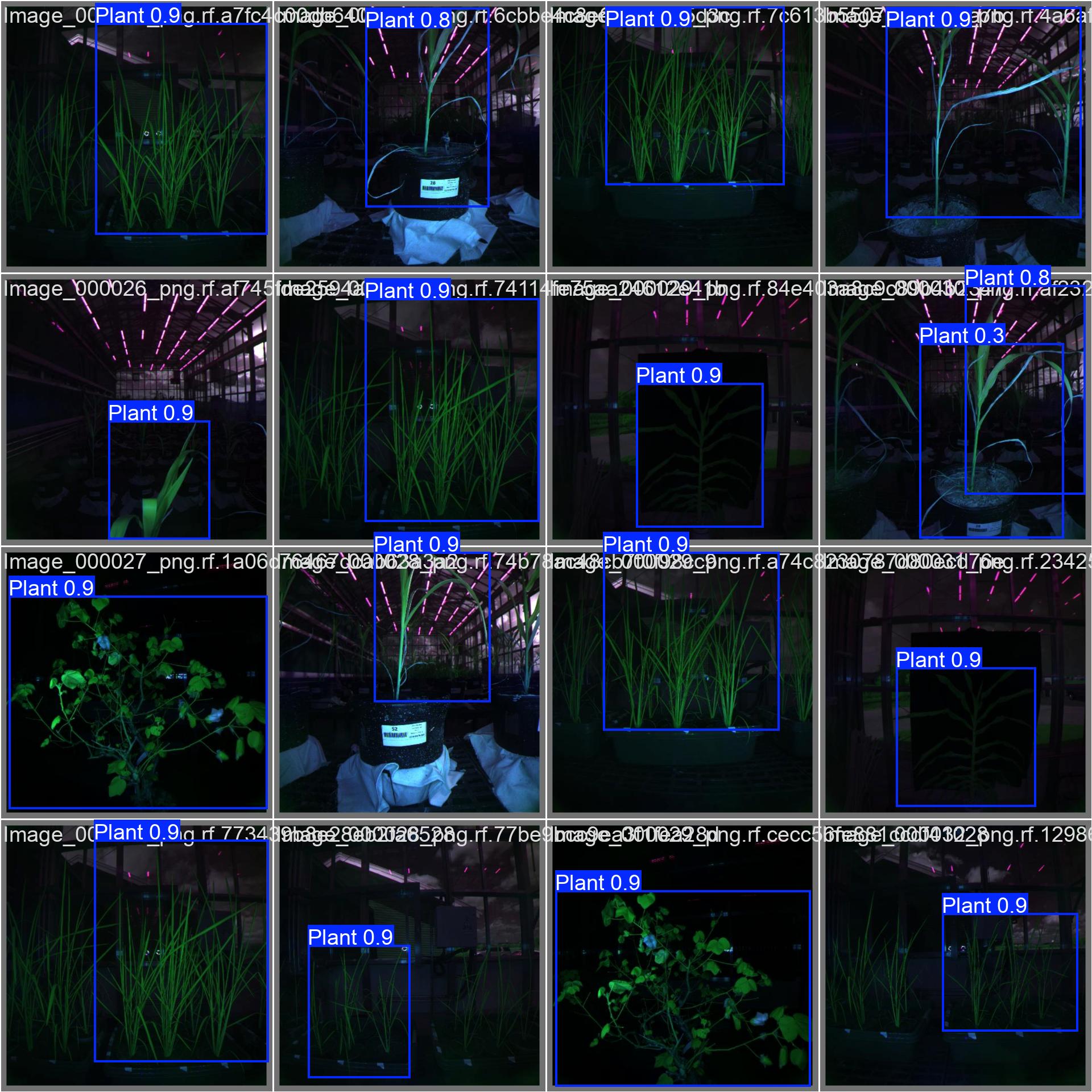}
    \caption{Sample predictions from a validation batch on the PGP dataset, illustrating detection accuracy and spatial localization.}
    \label{fig:pgp_validation_batch}
\end{figure}

\subsection{Benchmark Datasets Results}

\noindent\textit{MelonFlower Dataset Analysis} The MelonFlower dataset \cite{melon_flower_dataset}, sourced from Roboflow, features greenhouse-captured images and serves as an effective benchmark for evaluating the detection of small objects such as melon flowers across varied plant geometries. From Table~\ref{tab:combined_results}, it is evident that our proposed \textit{CBAM-STN-TPS-YOLO} model substantially outperforms both the baseline YOLO and intermediate STN-YOLO models, particularly in terms of precision and mAP. The inclusion of CBAM allows the model to better suppress irrelevant background features and focus on target regions, while the TPS module enables adaptive spatial transformations critical for capturing non-rigid deformations in foliage and flower structure. This is further supported by the qualitative example, where YOLO detects a melon flower with an Intersection over Union (IoU) of 0.50, whereas CBAM-STN-TPS-YOLO achieves a superior IoU of 0.60. The improved detection accuracy and spatial localization highlight the model's robustness in handling fine-grained agricultural object detection tasks. These findings suggest that combining attention and flexible spatial alignment not only enhances precision but also makes the model suitable for real-world deployment in resource-constrained agricultural monitoring systems.

\subsection{Statistical Analysis}

\noindent\textit{Statistical Significance Analysis.} To rigorously assess the reliability of the observed performance improvements, we conducted statistical significance testing between the STN-YOLO and CBAM-STN-TPS-YOLO models using the PGP dataset. For this analysis, we considered evaluation metrics across 100 training epochs, including Accuracy, Precision, Recall, and mean Average Precision (mAP). A paired t-test was performed for each metric to evaluate whether the differences in model performance were statistically significant across epochs. This test was selected because both models were evaluated under identical experimental conditions and dataset splits, ensuring the comparability of the results. 

The null hypothesis assumed no difference in the means of the performance metrics between the two models, while the alternative hypothesis postulated a significant difference due to the architectural enhancements in CBAM-STN-TPS-YOLO. The results indicated statistically significant improvements in all four metrics (p-values $<$ 0.05), strongly suggesting that the introduction of Thin-Plate Splines (TPS) and Convolutional Block Attention Modules (CBAM) into the baseline STN-YOLO architecture contributes to more effective spatial feature alignment and selective attention in agricultural object detection tasks. These findings validate that the observed metric gains are not incidental but are attributable to the integration of TPS and CBAM modules.

\section{Conclusion and Future Work}

This study introduced a novel object detection pipeline—CBAM-STN-TPS-YOLO—that enhances the traditional YOLO framework by integrating Spatial Transformer Networks (STNs), Thin-Plate Splines (TPS), and the Convolutional Block Attention Module (CBAM). The model was specifically designed to address challenges in plant image detection tasks, such as non-rigid deformations, cluttered backgrounds, and occlusions—scenarios commonly encountered in agricultural datasets. By leveraging TPS-based spatial transformations and CBAM for refined feature attention, our model significantly improves precision, recall, and mean average precision (mAP), while maintaining low inference time, making it suitable for real-time smart farming applications.

Experimental results across benchmark datasets (PGP and MelonFlower) demonstrated the superiority of CBAM-STN-TPS-YOLO over baseline models, with notable reductions in false positives and improved feature localization. The ablation studies further underscored the individual contributions of TPS and CBAM in enhancing spatial flexibility and attention-guided feature refinement.

Future directions for this work include: (1) extending the current pipeline to other modern detection frameworks (\textit{e.g.}, YOLOv10 \cite{yolov10}); (2) designing improved objective functions to further guide TPS-based deformation learning; (3) expanding the PGP dataset to encompass a wider variety of crops, growth stages, and environmental conditions including varied illumination; and (4) integrating multimodal data, such as near-infrared imagery, to enrich the spatial and spectral information for improved phenotypic feature extraction and analysis.

\section{Acknowledgment}
This material is based upon work supported by Texas A\&M AgriLife. Portions of this research were conducted with the advanced computing resources provided by Texas A\&M High Performance Research Computing (HPRC). 

This work builds upon the foundational contributions of Zambre \textit{et al.} \cite{zambre2024spatialtransformernetworkyolo}, whose integration of Spatial Transformer Networks (STNs) into the YOLO architecture inspired key components of our proposed CBAM-STN-TPS-YOLO pipeline.

\balance
\bibliographystyle{IEEEtran}
\bibliography{main.bib}

\end{document}